\documentclass{article}
\usepackage{spconf,amsmath,graphicx}
\usepackage{algorithm}
\usepackage{algorithmic}
\usepackage{hyperref}
\usepackage[english]{babel}
\usepackage{amsmath}
\usepackage{amssymb}
\usepackage{mathtools}
\usepackage{amsthm}
\usepackage[textsize=tiny]{todonotes}
\usepackage{multirow}
\usepackage{microtype}
\usepackage{subfigure}
\usepackage{booktabs}

\usepackage{newfloat}
\usepackage{listings}
\usepackage{bbding}
\usepackage{pifont}
\usepackage{amssymb}

\usepackage{enumitem}

\title{ViTASD: Robust Vision Transformer Baselines for Autism Spectrum Disorder Facial Diagnosis}
\name{Xu Cao$^{1,*}$, Wenqian Ye$^{1,*}$\thanks{*Equal contribution.}, Elena Sizikova$^{1}$, Xue Bai$^{2}$, Megan Coffee$^{3}$, Hongwu Zeng$^{2}$, Jianguo Cao$^{2,\dag}$ \thanks{This work was supported by Guangdong Province High-level Hospital Construction Project (Shenzhen Children's Hospital, 2021). Xu Cao and Wenqian Ye are IEEE Member.}}
\address{
    $^{1}$ New York University, New York, USA \\
    $^{2}$ Shenzhen Children's Hospital, Shenzhen, China \\
    $^{3}$ NYU Grossman School of Medicine, New York, USA \\
    \dag  Corresponding Author.
}
\begin{document}
%
\maketitle
\begin{abstract}

Autism spectrum disorder (ASD) is a lifelong neurodevelopmental disorder with very high prevalence around the world. Research progress in the field of ASD facial analysis in pediatric patients has been hindered due to a lack of well-established baselines. In this paper, we propose the use of the Vision Transformer (ViT) for the computational analysis of pediatric ASD. The presented model, known as ViTASD, distills knowledge from large facial expression datasets and offers model structure transferability. Specifically, ViTASD employs a vanilla ViT to extract features from patients' face images and adopts a lightweight decoder with a Gaussian Process layer to enhance the robustness for ASD analysis. Extensive experiments conducted on standard ASD facial analysis benchmarks show that our method outperforms all of the representative approaches in ASD facial analysis, while the ViTASD-L achieves a new state-of-the-art. Our code and pretrained models are available at \url{https://github.com/IrohXu/ViTASD}.

\end{abstract}

\section{Introduction}
\label{sec:intro}

Over the past decade, the significant clinical and scientific value of computer-assisted diagnosis (CAD) based on machine learning has been increasingly recognized. In particular, neural networks and transfer learning offer the benefit of learning compact and fixed dimensional representations from large-scale public datasets and utilizing the resulting representation to finetune models in various fields of medicine. Recent research shows that neural networks can be effective clinical aids for mental illness prevention~\cite{durstewitz2019deep}. However, to date, the progress of neural network approaches applied to the analysis of pediatric autism spectrum disorder (ASD) is limited, in part due to the fact that ASD is a heterogeneous neurodevelopmental disorder with complex cognitive features~\cite{cao2023commentary}. As a result, there are substantial difficulties in collecting ASD patient data and designing accurate CAD systems. Considering the high prevalence of ASD children, there is an urgent need for more robust ASD early diagnosis tools in clinical practice.

Quantifiable indices for ASD diagnosis have received much attention~\cite{cao2023commentary,de2020computer}. In previous studies, most neural network-based diagnoses of ASD focused on neuroimaging-based approaches~\cite{anagnostou2011review}. However, neuroimaging data collection is often challenging due to non-cooperation from pediatric patients~\cite{khodatars2021deep}. Recently, techniques based on behaviour analysis and affective computing have been introduced to address some of these issues. Their key idea is that patients with ASD exhibit altered attention and emotion to specific features of visual information~\cite{de2020computer}. For ASD children, changes are reflected in facial expressions and eye movement information~\cite{drimalla2020towards}. Compared with traditional neuroimaging methods, these new methods predict potential ASD risk by directly analyzing a patient's face, eye-tracking (ET) and behavior, all of which have the potential of integrating with other standardized assessments, such as the autism diagnostic observation schedule (ADOS).

\begin{figure*}[!t]
\centerline{\includegraphics[width=0.85\linewidth]{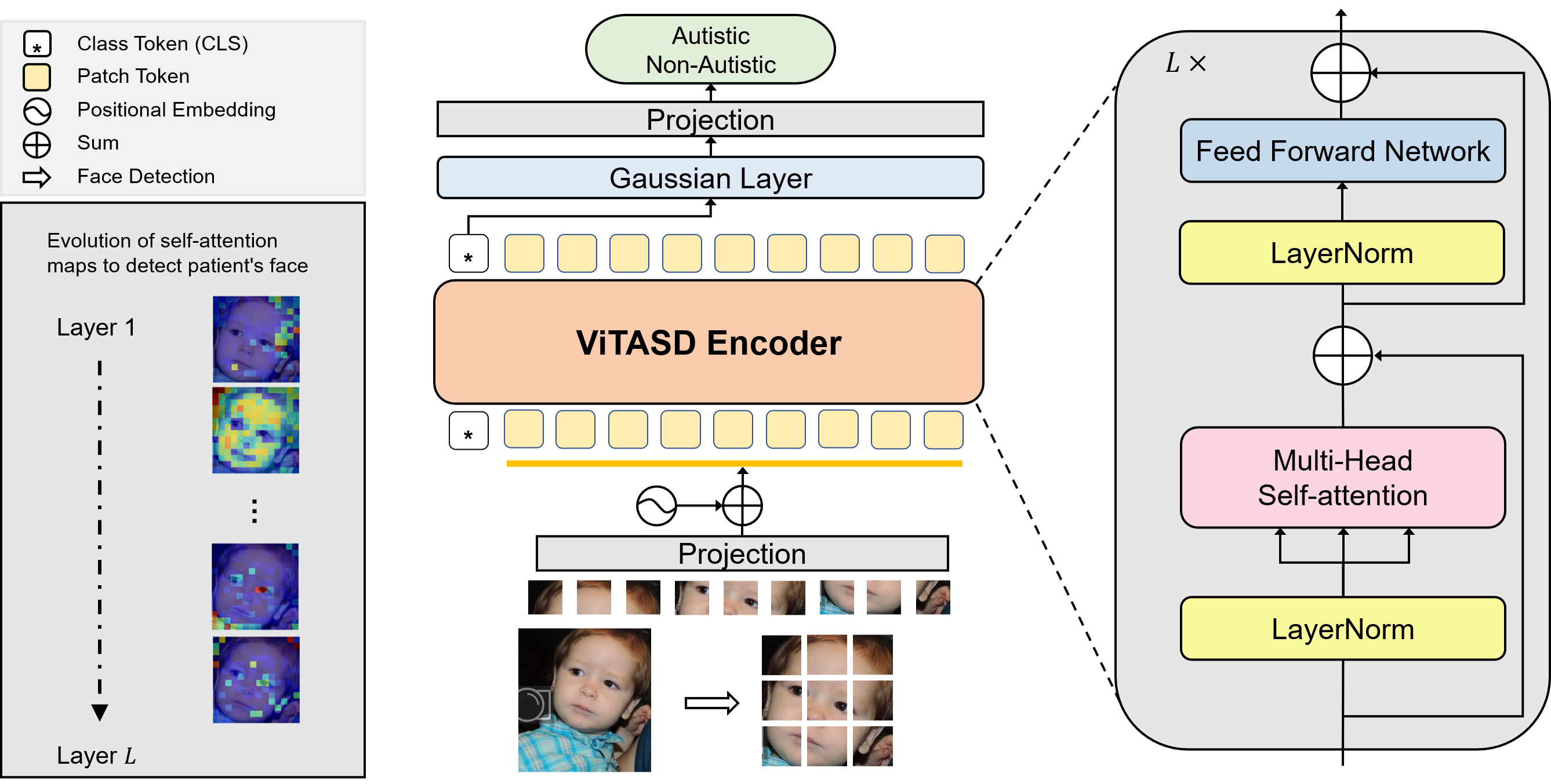}}
\caption{Overview of the components of the proposed ViTASD model for pediatric autism spectrum disorder (ASD) prediction from facial images. ViTASD consists of a Vision Transformer (ViT) encoder, a Gaussian layer-based decoder and a knowledge distillation (KD) module.}
\label{fig:framework}
\end{figure*}

Recent studies show that neural network-based methods can successfully distinguish between ASD and non-ASD children, given sufficient and well-annotated facial images~\cite{hosseini2021deep}. In this work, Hosseini et al. explored different convolutional neural networks (CNN) for facial analysis in ASD children and noted that some models, such as MobileNet~\cite{howard2017mobilenets}, can attain an accuracy of 90\% in classification of ASD and non-ASD patients. These results imply that neural networks can learn useful facial risk markers of ASD. In other works, Xie et al.~\cite{xie2022identifying} and Han et al.~\cite{han2022multimodal} designed neural network models using ET data to predict ASD. Their experiments demonstrated that image data (face, ET) collected in vitro is a very promising direction for designing ASD CAD systems. 

Existing facial analysis methods for ASD detection rely on CNNs for local feature extraction or as additional eye-tracking object detection modules. On the other hand, ViTs, in comparison to CNNs, better capture global context and are less biased toward local textures~\cite{naseer2021intriguing}. However, in small datasets, ViTs often perform worse than CNNs. This limitation motivates us to think from a new direction: how to utilize the strong representation learned from the pretrained ViT in ASD facial analysis? In this work, we propose ViTASD, a novel KD transformer baseline for ASD facial analysis. Our contributions can be summarized as follows.
\begin{itemize}[leftmargin=*,itemsep=0pt,topsep=0pt]
    \item We propose ViTASD, a novel model for automatic prediction of the autism spectrum disorder (ASD) in pediatric patients from facial images. ViTASD obtains state of the art performance on the autism spectrum disorder children's dataset~\cite{autismdataset}, a public benchmark.
    \item We empirically demonstrate capabilities of the ViTASD model: ability to scale model size, support of newest masked autoencoder (MAE) self-supervised learning, knowledge transferability from a large-scale facial expression dataset and large pretrained models.
\end{itemize}

\section{Methodology}
\label{sec:method}
An overview of our proposed framework can be seen in Fig.~\ref{fig:framework}. The goal of ViTASD is to provide a simple yet robust baseline for pediatric ASD detection from facial images. Thus, we aim to keep the original ViT structure~\cite{dosovitskiy2020image} without the addition of more complex modules. The decoder of ViTASD is a lightweight multilayer perceptron (MLP) layer with an optional Gaussian layer for Out-of-Distribution (OOD) data points.

\subsection{Structure of ViTASD baseline}
Given a patient face image $X \in R^{H \times W \times 3}$, ViTASD slices the input image into $16 \times 16$ patches via the patch embedding layer. The patches are then flattened to a $K \in R^{(\frac{H \times W}{16 \times 16} + 1) \times D}$ output with an additional class token. Here, $D$ is the channel dimension. Finally, the tokens are processed by several Transformer blocks, each composed of a multi-head self-attention layer (MHA) and a multi-layer perceptron layer (MLP):

\begin{align}
    K_{i+1}' = MHA_{i+1}((LN(K_i)) + K_{i} \\
    K_{i+1} = MLP_{i+1}(LN(K_{i+1}')) + K_{i+1}',
\end{align}

\noindent where $i$ is the output of $i$-th Transformer block; $MHA_{i+1}$ is the multi-head self-attention layer of the $i+1$-th Transformer block; $MLP_{i+1}$ is the multi-layer perceptron of the $i+1$-th Transformer block.

\begin{figure}[!t]
\centerline{\includegraphics[width=0.9\linewidth]{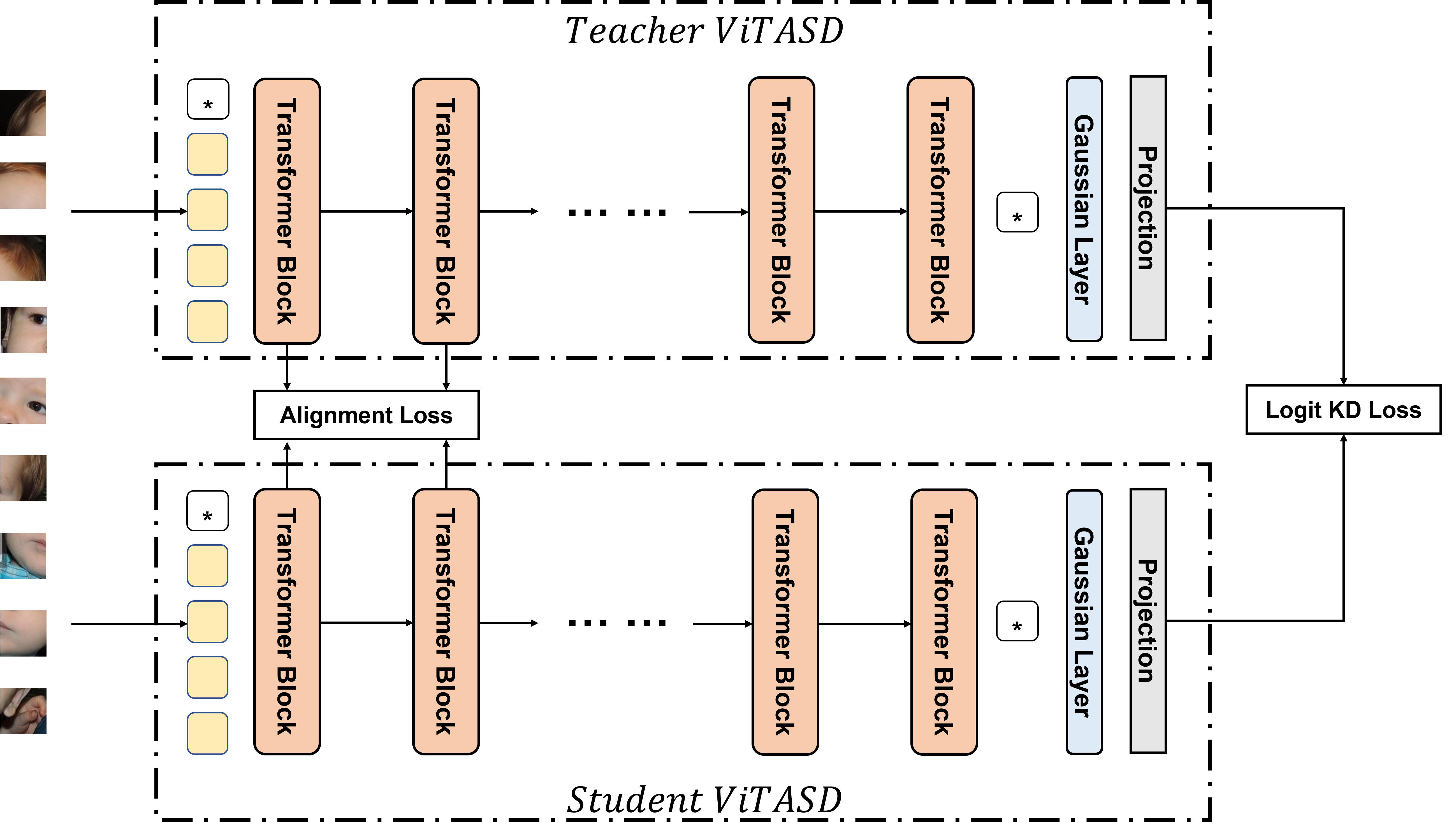}}
\caption{Knowledge distillation of ViTASD.}
\label{fig:vitkd}
\end{figure}

\subsection{Properties of ViTASD}
\noindent\textbf{Pretraining data flexibility.} In contrast to CNN-based methods, ViTASD benefits from the data flexibility from ViT in both transfer learning and representation learning. In our model, we explore data flexibility other than the default settings of ImageNet-21k pretraining. In the ablation study (see Sec.~\ref{sec:ablation}), we prove that both supervised and self-supervised pretraining using AffectNet can improve the ViTASD's performance.

\noindent\textbf{Model structure transferability.} 
We empirically demonstrate that ViTASD can distill knowledge from larger structures to match performance in smaller ones. To fill the structure gap, we introduce two distillation losses, the alignment loss $\mathcal{L}_{align}$ and the logit KD Loss $\mathcal{L}_{logit}$, to substantially improve performance of the student network (see Fig. \ref{fig:vitkd}). $\mathcal{L}_{align}$ aligns the feature distillation on the attention maps of the shallow layers (e.g., layers 0 and 1) using Mean Square Error (MSE), where $\text{FC} (\cdot)$ is a linear layer to reshape the $\mathcal{F}^{Teacher}$ to the same dimension as $\mathcal{F}^{Student}$. $\mathcal{L}_{logit}$ ensures the classification logits of the student ViTASD match those of the teacher ViTASD. To summarize, we train the student model with the following total loss:
\begin{equation} \label{eq:KD}
\begin{aligned}
&\mathcal{L}_{align} = \text{MSE}(\text{FC} (\mathcal{F}^{Teacher}) - \mathcal{F}^{Student}) \\
&\mathcal{L}_{logit} = \text{MSE}(\text{logit}^{Teacher}(x), \text{logit}^{Student}(x)) \\
&\mathcal{L}_{KD} =\mathcal{L}_{logit}+\alpha \mathcal{L}_{align},
\end{aligned}
\end{equation}
where $\alpha$ is a hyperparameter to balance the distillation loss.

\noindent\textbf{Finetuning flexibility.}
We add a Gaussian Process Layer decoder with an RBF kernel to ViTASD to enable finetuning on out-of-domain data. This modification improves the uncertainty representation of the model, hence enhancing its robustness. This layer is implemented as a two-layer network:

\begin{equation}
    \operatorname{logits}(x)=\Phi(x) \beta, \quad \Phi(x)=\sqrt{\frac{2}{M}} * \cos (W x+b)
\end{equation}
where $x$ is the input, and $W$ and $b$ are frozen weights initialized randomly from Gaussian and uniform distributions, respectively. $\Phi(x)$ are the Random Fourier Features (RFF) \cite{williams2006gaussian}. $\beta$ is a learnable kernel weight similar to that of a Dense layer.




\section{Experiments}
\label{sec:experiments}

\subsection{Implementation details}
ViTASD follows the DeiT III~\cite{touvron2022deit} architecture and the OOD task setting~\cite{fort2021exploring} for ViT, i.e., three augmentations (grayscale, solarization, Gaussian blur) and Cut-Mix, Mix-Up. We use ViT-S~\cite{dosovitskiy2020image}, ViT-B~\cite{dosovitskiy2020image}, and ViT-L~\cite{dosovitskiy2020image} as encoder backbones and denote the corresponding models as ViTASD-S, ViTASD-B, ViTASD-L. The backbones are initialized with two settings: (a) MAE~\cite{he2022masked} pretrained weights from AffectNet~\cite{mollahosseini2017affectnet} (the largest facial expressions database); (b) general supervised learning pretrained weights from AffectNet~\cite{mollahosseini2017affectnet}. We use the 224$\times$224 input resolution and AdamW optimizer with a learning rate of 1e-4. The $\alpha$ for KD training is 5e-5. All models are trained for 300 epochs with batch size of 128 on 4 NVIDIA A100 GPUs.

\begin{figure*}[!t]
\centerline{\includegraphics[width=0.80\linewidth]{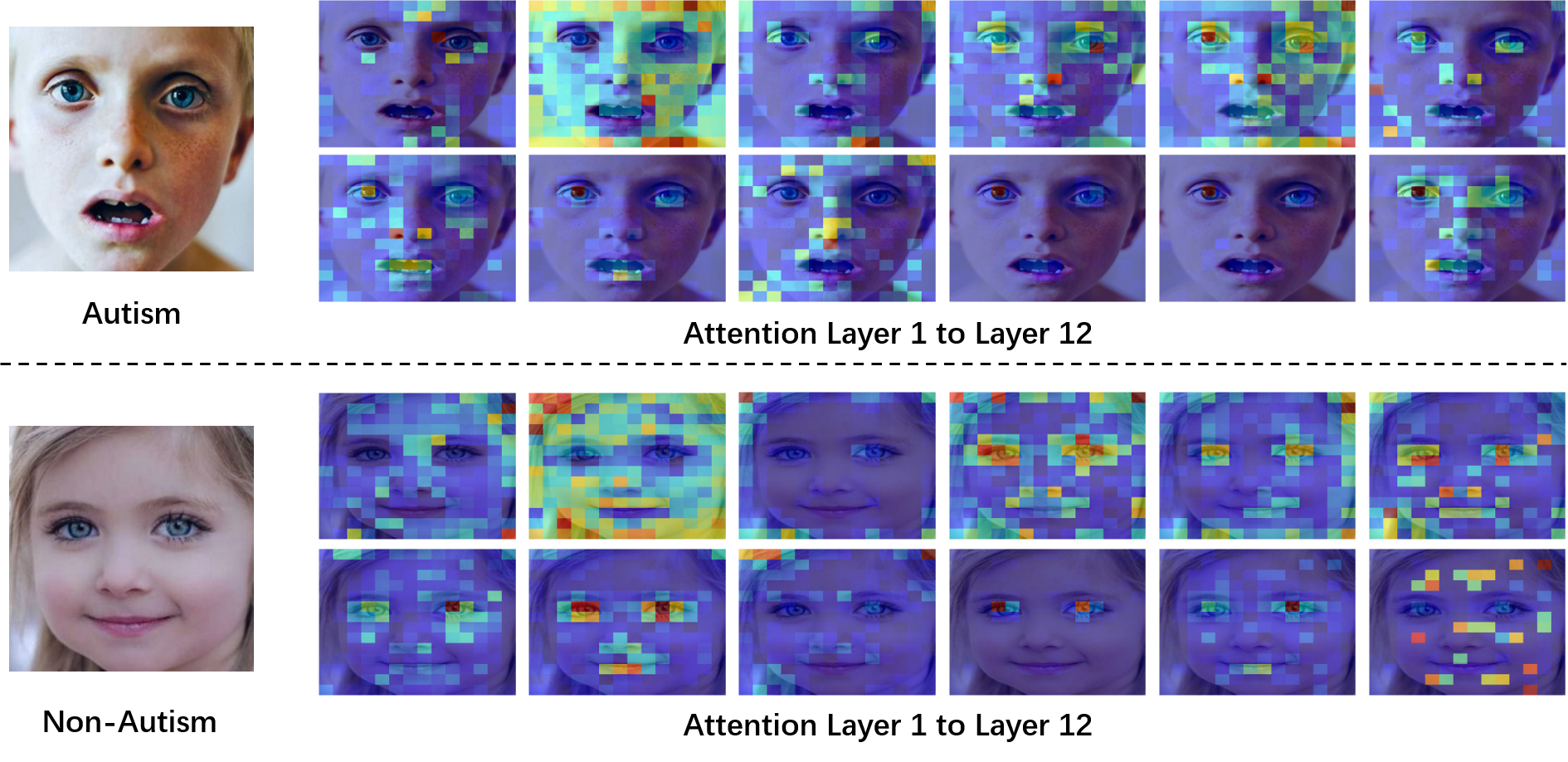}}
\caption{Visualization of the attention maps between the ASD classification token and all visual tokens in the 12 transformer layers of ViTASD-B, where {\color{red}red} and {\color{blue}blue} represent regions with high and low attention, respectively.}
\label{fig:attention}
\end{figure*}

\begin{table}
\begin{center}
\caption{The performance of ViTASD in different model scales and pretrained settings.}
\small
\label{table:ab1}
\begin{tabular}{l|c|c|c}
\toprule
Methods & Params & Pretrained & Accuracy $\uparrow$ \\
\midrule

ResNet50 & 23.5M & ImageNet-21k & 91.00 $\pm$ 0.24 \\
ResNet152 & 60.3M & ImageNet-21k & 91.33 $\pm$ 0.24 \\
ResNet152 & 60.3M & AffectNet~\cite{mollahosseini2017affectnet} & 89.50 $\pm$ 0.24 \\
\midrule
ViTASD-S  & 27.1M & ImageNet-21k & 91.00 $\pm$ 0.41  \\
ViTASD-B & 85.8M &  ImageNet-21k & 92.83 $\pm$ 0.24 \\
ViTASD-L & 307M &  ImageNet-21k & 93.17 $\pm$ 0.24  \\
ViTASD-L & 307M &  MAE~\cite{he2022masked} & 94.00 $\pm$ 0.41  \\
ViTASD-L & 307M &  AffectNet~\cite{mollahosseini2017affectnet}  & \textbf{94.50 $\pm$ 0.23}  \\

\bottomrule
\end{tabular}
\end{center}
\end{table}

\begin{table}
\begin{center}
\caption{Knowledge distillation effect on ViTASD.}
\small
\label{table:ab2}
\begin{tabular}{l|l|cc}
\toprule
Student & Teacher & Teacher Pretrained & Accuracy $\uparrow$ \\
\midrule
ViTASD-S & ViTASD-B & AffectNet~\cite{mollahosseini2017affectnet} & 93.50 $\pm$ 0.24  \\
ViTASD-B & ViTASD-L & AffectNet~\cite{mollahosseini2017affectnet} & \textbf{94.00 $\pm$ 0.24}  \\
\bottomrule
\end{tabular}
\end{center}
\end{table}


\begin{table*}
\begin{center}
\caption{Comparison of ViTASD and SOTA methods on the test set of the autism spectrum disorder (ASD) children's dataset~\cite{autismdataset}.}
\label{table:sota}
\begin{tabular}{l|c|c|c|ccc}
\toprule
Methods & Backbone & Params & Pretrained & Accuracy$\uparrow$ &  AUROC$\uparrow$ \\
\midrule
\cite{alsaade2022classification,elshoky2022comparing} & VGG-19 & 139M &  ImageNet-21k & 90.50 $\pm$ 0.41 & 93.65 $\pm$ 0.13 \\
\cite{elshoky2022comparing} & ResNet50 & 23.5M & ImageNet-21k & 91.00 $\pm$ 0.23 & 94.82 $\pm$ 0.62 \\
\cite{alsaade2022classification,ahmed2022facial,hosseini2021deep} & MobileNetV3 & 4.2M & ImageNet-21k & 91.00 $\pm$ 0.23 & 94.43 $\pm$ 0.35\\
\cite{mujeeb2022identification} & EfficientNet-B4 & 17.6M & ImageNet-21k & 91.00 $\pm$ 0.41 & 95.13 $\pm$ 0.26\\
\cite{alsaade2022classification} & Xception & 20.8M & ImageNet-21k & 91.33 $\pm$ 0.24 & 95.40 $\pm$ 0.16\\
\midrule
ViTASD-B & ViT-B & 85.8M & ImageNet-21k & 92.83 $\pm$ 0.24 & 96.94 $\pm$ 0.10 \\
ViTASD-B & ViT-B + knowledge distillation & 85.8M &  AffectNet~\cite{mollahosseini2017affectnet} & 94.00 $\pm$ 0.24 & 97.16 $\pm$ 0.48\\
ViTASD-L & ViT-L & 307M & AffectNet~\cite{mollahosseini2017affectnet} & \textbf{94.50 $\pm$ 0.23} & \textbf{97.92 $\pm$ 0.12} \\

\bottomrule
\end{tabular}
\end{center}
\end{table*}

We use the largest publicly available facial expression recognition dataset AffectNet~\cite{mollahosseini2017affectnet} (a facial expression dataset with more than 1M images) to pretrain ViTASD in both supervised and self-supervised ways. For self-supervised learning, we adopt MAE~\cite{he2022masked} by randomly masking 75\% patches from the input images and reconstructing those masked patches. The performance of ViTASD is evaluated on the Autism spectrum disorder (ASD) children's dataset~\cite{autismdataset}, which consists of 2,926 images of resolution 224$\times$224 pixels with binary labels (non-autism and autism). The dataset is split into training set (2,526), validation set (200), and test set (200) in its updated version (see \cite{autismdataset} for more details). We measure performance using accuracy and area under the receiver operating characteristic (AUROC) on binary labels (non-autism and autism). 


\subsection{Quantitative evaluation} \label{sec:ablation}
We report performance comparisons between ViTASD and the state-of-the-art approaches are shown in Table~\ref{table:sota}. From the results, we make the following observations. (i) A pre-trained ViT is significantly better in accuracy and AUROC metrics than any CNN-based methods for ASD facial detection. (ii) The representations learned from the large-scale facial expression dataset (AffectNet) are helpful for transfer learning in ASD. With AffectNet pretraining, the performance of ViTASD-L further increases to 94.50 accuracy, implying the good knowledge transferability and flexibility of ViTASD.

We also evaluate ViTASD with different ViT backbones and report results in Table~\ref{table:ab1}. We find that accuracy improvement between ResNet-50 and ResNet-152 is significantly lower than between ViTASD-S and ViTASD-B, demonstrating that the ViT model can better learn representation via large facial dataset and transfer into a new ASD facial analysis task. In Table~\ref{table:ab2}, we further investigate the KD performance on ViTASD-B to ViTASD-S and ViTASD-L to ViTASD-B with performance loss of $0.5\%$. The result illustrates the strong model structure transferability of ViTASD.


 
\subsection{Visualization and interpretability}
In order to show the interpretability of the proposed ViTASD, we visualize the attention maps during inference on the test set in Fig.~\ref{fig:attention}. The attention map is the interaction between the classification token and all visual tokens. The attention scores, which showed the color in the attention maps, can be used to understand which areas contribute most to the classification result. For both autism and non-autism children's' face images, the model attends most to the eye region, which is known to be one of the most distinguishable features of the autism children in clinical practice~\cite{kwon2019typical}.

\section{Conclusion and Discussion}
\label{sec:conclusion}

In this paper, we propose ViTASD, the first ViT-based baseline for pediatric ASD diagnosis. We have shown that pediatric ASD can be formulated as a facial image classification problem using a ViT, which achieves state-of-the-art performance in both accuracy and AUROC, while generating attention maps consistent with distinguishable ASD features. We hope this work could provide insights to the biomedical imaging and signal processing community for ASD research and inspire further study on exploring the potential of applying explainable ViTs in more facial analysis application tasks.







\bibliographystyle{IEEEbib}
\bibliography{ref}

\end{document}